\documentclass[letterpaper, 10 pt, conference]{ieeeconf}  

\IEEEoverridecommandlockouts                              

\usepackage{cite}
\usepackage{amsmath,amssymb,amsfonts}
\usepackage{algorithmic}
\usepackage{graphicx}
\usepackage{textcomp}
\usepackage{xcolor}
\usepackage{enumerate}
\usepackage[linesnumbered,ruled,lined]{algorithm2e}
\usepackage{booktabs}
\usepackage{threeparttable}
\usepackage{multirow}
\usepackage{stfloats}
\usepackage{subfigure}
\usepackage{graphicx}
\usepackage{rotating}
\usepackage{geometry}
\title{\LARGE \bf
CFP-SLAM: A Real-time Visual SLAM Based on Coarse-to-Fine Probability in Dynamic Environments
}

\author{Xinggang Hu$^{1}$, Yunzhou Zhang$^{1*}$, Zhenzhong Cao$^{1}$,   \\Rong Ma$^{2}$, Yanmin Wu$^{3}$, Zhiqiang Deng$^{1}$, Wenkai Sun$^{1}$ 
	\thanks{$^*$The corresponding author of this paper. }
	\thanks{$^{1}$Xinggang Hu, Yunzhou Zhang, Zhenzhong Cao, Zhiqiang Deng and Wenkai Sun are with College of Information Science and Engineering, Northeastern University, Shenyang 110819, China (Email: {\tt\small zhangyunzhou@mail.neu.edu.cn}).}%
	\thanks{$^{2}$Rong Ma is with Beijing Simulation Center, China (Email: {\tt\small mar\_buaa@163.com}).}
	\thanks{$^{3}$Yanmin Wu is with School of Electronic and Computer Engineering, Peking University, Shenzhen, China.}%
	\thanks{This work was supported by National Natural Science Foundation of China (No. 61973066), Major Science and Technology Projects of Liaoning Province(No.2021JH1/10400049), Fundation of Key Laboratory of Equipment Reliability(No.WD2C20205500306), Fundation of Key Laboratory of Aerospace System Simulation(No.6142002200301).}
}

\begin{document}

\maketitle
\thispagestyle{empty}
\pagestyle{empty}

\begin{abstract}
The dynamic factors in the environment will lead to the decline of camera localization accuracy due to the violation of the static environment assumption of SLAM algorithm. Recently, some related works generally use the combination of semantic constraints and geometric constraints to deal with dynamic objects, but problems can still be raised, such as poor real-time performance, easy to treat people as rigid bodies, and poor performance in low dynamic scenes. In this paper, a dynamic scene-oriented visual SLAM algorithm based on object detection and coarse-to-fine static probability named CFP-SLAM is proposed. The algorithm combines semantic constraints and geometric constraints to calculate the static probability of objects, keypoints and map points, and takes them as weights to participate in camera pose estimation. Extensive evaluations show that our approach can achieve almost the best results in high dynamic and low dynamic scenarios compared to the state-of-the-art dynamic SLAM methods, and shows quite high real-time ability.
\end{abstract}

\section{INTRODUCTION}

Simultaneous localization and mapping (SLAM) is the key technology for autonomous navigation of mobile robots, and it is widely applied in the fields of autopilot, UAV and augmented reality (AR). SLAM system is based on environmental static assumption\cite{saputra2018visual}, and dynamic factors will bring wrong observation data to the system, making it difficult to establish various geometric constraints on which SLAM system works, and reducing the accuracy and robustness of SLAM system. The abnormal point processing mechanism of RANSAC (Random Sample Consensus) algorithm can solve the influence of certain abnormal points in static or slightly dynamic environment. However, when dynamic objects occupy most of the camera view, RANSAC algorithm has little effect.
 
With the development of deep learning technology, some advanced researchers have used semantic constraints to solve the visual SLAM problem in dynamic environment recent years. The general approach is to take the semantic information obtained from object detection \cite{zhong2018detect,xiao2019dynamic} or semantic segmentation \cite{yu2018ds,bescos2018dynaslam,brasch2018semantic,wang2019unified,yuan2020sad,vincent2020dynamic,li2021dp,ji2021towards,fan2022blitz} as a priori and eliminate the dynamic objects in the environment combined with geometric constraints. Semantic segmentation can provide a fine pixel level object mask, but its real-time performance is poor. The improvement of segmentation accuracy and robustness often comes at the cost of huge computational cost. Even so, the segmentation boundary of the object can not be extremely accurate and can not completely cover the moving object\cite{fan2022blitz}. Object detection can circumvent the problems above, but there are a large amount of background point clouds in the box of objects, and some complex cases will be missed easily\cite{xiao2019dynamic}. In addition, there are two common problems with current schemes: 1) All dynamic objects are treated as high dynamic attributes, which leads to poor performance in low dynamic scene. 2) As non-rigid objects, human bodies often perform partial movement. Directly eliminating the human body as a whole object will reduce the constraint of keypoints and introduce a negative effect on accuracy of localization.


For the above problems, we propose CFP-SLAM, which is a high-performance high-efficiency visual SLAM system based on object detection and static probability in indoor dynamic environments. On the basis of ORB-SLAM2\cite{mur2017orb}, CFP-SLAM uses YOLOv5 to obtain semantic information, uses extended Kalman filter (EKF) and Hungarian algorithm to compensate missed detection, calculates the static probability of objects to distinguish high dynamic objects from low dynamic objects, and distinguishes foreground points and background points of object detection results based on DBSCAN (Density-Based Spatial Clustering of Applications with Noise) algorithm. Established on a variety of constraints, a two-stage calculation method of the static probability of keypoints from coarse to fine is designed. The static probability of keypoints is used as a weight to participate in the camera pose optimization. Considering the needs of different scenarios, we provide a lower-performance version to improve the real-time performance without calculating the static probability of objects.

Extensive experiments are conducted on public datasets. Compared with state-of-the-art dynamic SLAM methods, our approach achieves the highest localization accuracy in almost all low dynamics and high dynamic scenarios. \textbf{The main contributions of this paper are as follows:}
\begin{itemize}
	\item Compensating missed detection based on EKF and Hungarian algorithm, while using DBSCAN clustering algorithm to distinguish the foreground points and background points of box.
	\item The distinction of object dynamic attributes. Based on the YOLOv5 object detection and geometric constraints, the object motion attributes are divided into high dynamics and low dynamics, which are provided to the subsequent methods as a priori information for processing with different strategies, so as to improve the robustness and adaptability of SLAM system.
	\item The static probability of keypoints from coarse to fine. A two-stage static probability of keypoints calculation method based on the static probability of object, the DBSCAN clustering algorithm, the epipolar constraints and the projection constraints is proposed to solve the problem of false deletion of static keypoints caused by non-rigid body local motion.
\end{itemize} 



\section{Related Work}
\subsection{Dynamic SLAM without Priori Semantic Information}

When there is no semantic information as the priori, using reliable constraints to find the correct feature matching relationship is the basic method to deal with dynamic SLAM problem. Li et al.\cite{li2017rgb} propose a static weighting method of keyframe edge points, and integrated into the IAICP method to reduce tracking error. Sun et al. \cite{sun2017improving} roughly detect the motion of moving objects based on self motion compensation image difference, and enhance the motion detection by tracking the motion using particle filter. Then, they \cite{sun2018motion} propose a novel RGB-D data-based on-line motion removal approach, and build and update the foreground model incrementally. StaticFusion \cite{scona2018staticfusion} simultaneously estimates the camera motion as well as a probabilistic static/dynamic segmentation of the current RGB-D image pair. DMS-SLAM \cite{liu2019dms} uses GMS\cite{bian2017gms} to eliminate mismatched points. Kim et al.\cite{kim2016effective} propose a dense visual mileage calculation method based on background model to estimate the nonparametric background model from depth scene. Dai et al.\cite{dai2020rgb} distinguishe dynamic and static map points based on feature correlation. Flowfusion \cite{zhang2020flowfusion} uses optical flow residuals to highlight dynamic regions in rgbd point clouds. Because there is no need for deep learning networks to provide semantic priors, the above methods are usually fast in dealing with dynamic factors, but lack of accuracy.

\subsection{Dynamic SLAM Based on Semantic Constraints}

Semantic segmentation or object detection can provide a steady and reliable priority constraint for dynamic SLAM. Detect-SLAM \cite{zhong2018detect} detects objects in keyframes and propagates the motion probability of keypoints in real time to eliminate the influence of dynamic objects in SLAM. DS-SLAM \cite{yu2018ds} uses SegNet\cite{badrinarayanan2017segnet} to obtain semantic information, combines sparse optical flow and motion consistency detection to judge people's dynamic and static attributes. Dyna-SLAM \cite{bescos2018dynaslam} combines mask R-CNN \cite{he2017mask} and multi view geometry to process moving objects. Brasch et al.\cite{brasch2018semantic} present monocular SLAM approach for highly dynamic environments which models dynamic outliers with a joint probabilistic model based on semantic prior information predicted by a CNN. With the help of the initial segmentation results, Wang et al.\cite{wang2019unified} extract the accurate pose from the rough pose by identifying and processing the moving object and possible moving object respectively, and further help to make up for the error and boundary inaccuracy of the segmentation area. Dynamic-SLAM \cite{xiao2019dynamic} compensates SSD for missed detection based on the speed invariance of adjacent frames, and eliminates dynamic objects combined with selective tracking algorithm. SaD-SLAM \cite{yuan2020sad} extracts static feature points from objects judged as dynamic based on semantic by verifying whether the inter frame feature points meet the epipolar constraints. Vincent et al.\cite{vincent2020dynamic} perform semantic segmentation of object instances in the image, and use EKF to identify, track and remove dynamic objects from the scene. DP-SLAM \cite{li2021dp} combines the results of geometric constraints and semantic segmentation, the dynamic keypoints are tracked in the Bayesian probability estimation framework. Ji et al. \cite{ji2021towards} only perform semantic segmentation on keyframes, cluster the depth map and identifies moving objects combined with re-projection error to remove known and unknown dynamic objects. Blitz-SLAM \cite{fan2022blitz} repairs the mask of BlitzNet\cite{dvornik2017blitznet} based on depth information, and classifies static and dynamic matching points in potential dynamic areas using epipolar constraints. Generally, the above methods can accurately eliminate dynamic objects in the environment, but it is difficult to give consideration to both localization accuracy and real-time, and the performance is generally poor in low dynamic scenes.

\section{System Overview}
\subsection{Definition of Variables}
In this paper, common variables are defined as follows:
\begin{itemize}
\item $F_{k}$ - Frame K.

\item $K$ - The intrinsic matrix of a pinhole camera model.

\item $T_{k,w}\in{R^{4\times4}}$ - The transformation from world frame to camera frame K, which is composed of a rotation $R_{k,w}\in{R^{3\times3}}$ and a translation $t_{k,w}\in{R^{3\times1}}$.


\item $P_{i}^{k}$ -  The keypoint with ID $i$ in $F_{k}$. Its pixel coordinate is $P_{i_{u v}}^{k}=\left[u_{i}^{k}, v_{i}^{k}\right]^{T}$, camera coordinate is $P_{i_{k}}^{k}=\left[X_{i_{k}}^{k}, Y_{i_{k}}^{k}, Z_{i_{k}}^{k}\right]^{T}$, world coordinate is $P_{i_{w}}^{k}=\left[X_{i_{w}}^{k}, Y_{i_{w}}^{k}, Z_{i_{w}}^{k}\right]^{T}$. $\widetilde{(\cdot)}$ is the form of homogeneous coordinates in each coordinate system.

\item $P_{i^{*}}^{k-1}$ - The keypoint with ID $i^{*}$ in $F_{k-1}$ which forms a matching relationship with $P_{i}^{k}$.


\item $O_{i^{+}}^{k}$ - The static probability of potential moving object with ID $i^{+}$. $P_{i}^{k}$ is the extracted keypoint on the object.

\item $O_{T h}$ -  The threshold to distinguish whether the object motion attribute is high dynamic or low dynamic.

\item $K_{i}^{k}$ - The static probability of $P_{i}^{k}$, which is in the update state and participates in camera pose optimization.

\item $K _{i}^{D k}, K_{i}^{T k}, K_{i}^{F k}$ - The static probability of $P_{i}^{k}$ obtained by the DBSCAN clustering algorithm, the projection constraints and the epipolar constraints respectively. 


\item $M_{i^{-}}^{k}$ - The static probability of the map point forming a matching relationship with $P_{i}^{k}$.
\end{itemize}

 \begin{figure*}[]
\centering
 	\includegraphics[scale=0.57]{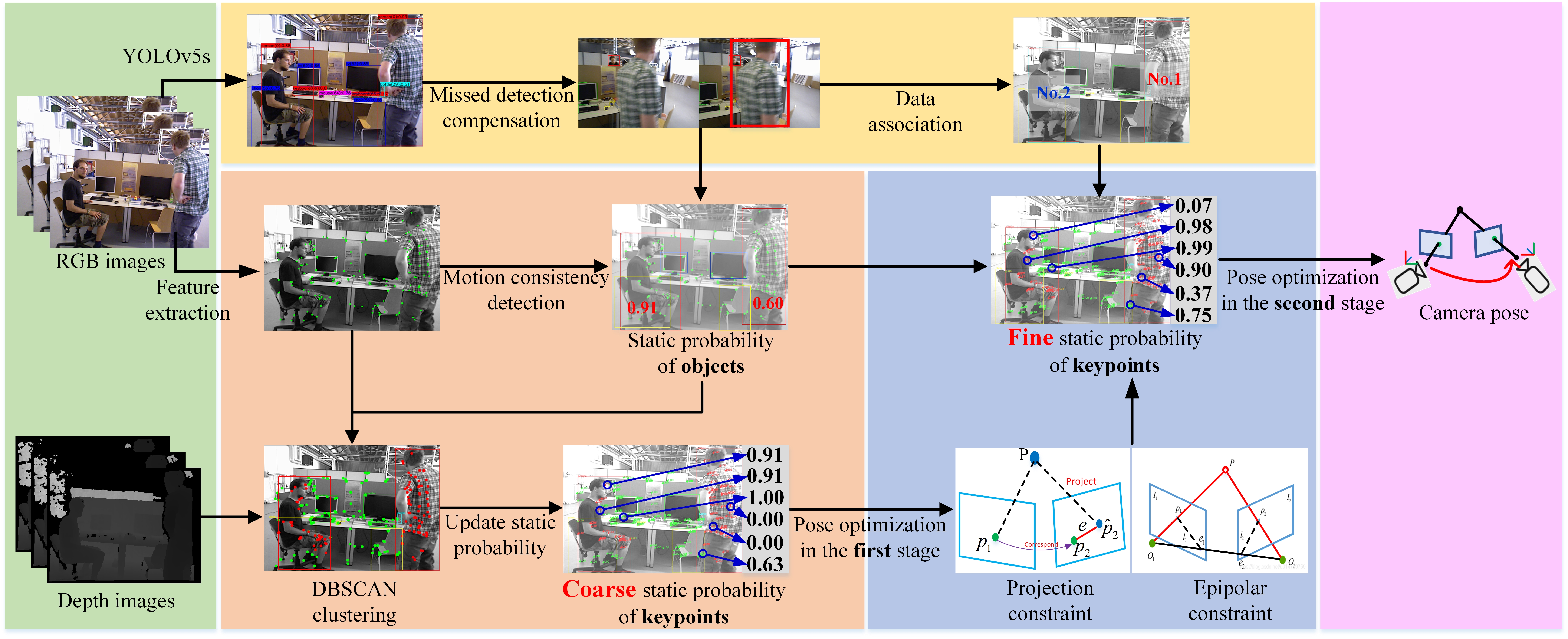}
 	\setlength{\abovecaptionskip}{-0.05cm} 
 	\caption{The overview of CFP-SLAM. The green portion and the purple portion are the input and output modules of the system respectively. The yellow portion is the semantic module, including object detection, missed detection compensation, and data association. The orange portion and the blue portion are static probability calculation modules for two stages of keypoints, respectively. In the first stage, the rough static probability of keypoints is calculated based on the static probability of objects and the results of DBSCAN clustering. In the second stage, based on the epipolar constraint and projection constraint, and considering the static probability of the object and the data association result of the box, the accurate static probability of feature points is calculated. During the whole process, the static probability of the map points is maintained and updated, and together with the static probability of the keypoints will be used as weight to participate in pose optimization.}
 	\label{Overview}
 	\vspace{-3.5mm} 
 \end{figure*}
 
 
\subsection{System Architecture}
The overview of CFP-SLAM is demonstrated in Fig.\ref{Overview}. Based on ORB-SLAM2\cite{mur2017orb}, we design a complete static probability calculation and update framework of keypoints based on multiple constraints to deal with the influence of moving objects in dynamic environment. The system obtains semantic information based on YOLOv5, compensates for missed detection based on EKF and Hungarian algorithm, and then the box between adjacent frames is associated. In $F_{k}$, only calculate and update the static probability of the keypoints inside the potential moving object box. Firstly, the static probability of potential moving object $O_{i^{+}}^{k}$ is obtained by using the optical flow and the epipolar constraints, and the object is divided into high dynamic object and low dynamic object. Initialize $K_{i}^{k}$ as the static probability of the object to which the keypoint belongs. Then, foreground points and background points is distinguished and the $K _{i}^{D k}$ is calculated by using the DBSCAN clustering results, and the $K_{i}^{k}$ is updated to estimate the camera pose in the first stage to obtain $T_{k,w}$. Next, $K_{i}^{T k}, K_{i}^{F k}$ are obtained by using the projection constraints and the epipolar constraints, $K_{i}^{k}$ and $M_{i^{-}}^{k}$ are updated to participate in camera pose optimization as weights to obtain a more accurate $T_{k,w}$.

\section{Specific implementation} 
\subsection{Missed Detection Compensation Algorithm}

When processing dynamic objects, if the semantic information as a priori is suddenly missing in some frames, on the one hand, the subsequent methods based on semantic priors will not be able to process dynamic objects. On the other hand, the sudden emergence of dynamic objects in high dynamic scenes will lead to a sharp increase in the number of keypoints incorrectly matched between adjacent frames, which leads to the loss of tracking in SLAM system in high dynamic scenario. Therefore, stable and accurate semantic information is critical.

In order to solve the missed detection problem of YOLOv5, we introduce EKF and Hungarian algorithm to compensate the missed detection of potential moving objects. EKF is used to predict the boxes of potential moving objects in $F_{k}$, while the Hungarian algorithm is used to correlate the predicted boxes with the boxes detected by YOLOv5. If the predicted box does not find a matching detected box, it could be considered that $F_{k}$ has missed detection, and the prediction result of EKF is adopted to compensate the missed detection result. After missed detection compensation, EKF and Hungarian algorithm are used again for inter frame data association of boxes. 

\subsection{Static Probability of Objects}

When calculating the static probability of each potential moving object, we use the idea of DS-SLAM \cite{yu2018ds} for reference to solve the fundamental matrix $L F_{k, k-1}$ and get the polar error $L d_{i}^{F_{k, k-1}}$. We use the epipolar constraints and chi-square distribution to test the epipolar error. Since the pixel coordinates of the matching point pair obtained by the optical flow tracking have $k=2$ degrees of freedom, if they are assumed to follow the Gauss Distribution $N(0,1)$, then according to the chi-square distribution:

\begin{equation}
\operatorname{chsq}(x ; k)=\left\{\begin{array}{c}
\frac{x^{(k / 2-1)} e^{-x / 2}}{2^{k / 2} \Gamma\left(\frac{k}{2}\right)}, x>0 \\
0, x \leq 0
\end{array}\right.
    \label{eq2}
\end{equation}

The definition of the function $\Gamma(v)$ is:

\vspace{-2mm}

\begin{equation}
\Gamma(v)=\int_{0}^{\infty} e^{-t} t^{v-1} d t,  \operatorname{Re} v>0
    \label{eq3}
\end{equation}

The single estimation result of $O_{i^{+}}^{k}$ can be obtained:

\begin{equation}
\left(O_{i^{+}}^{k}\right)_{m}=\operatorname{chsq}\left(\left(L d_{i}^{F_{k, k-1}}\right)^{2} ; 2\right)
    \label{eq4}
\end{equation}

After all estimation results are obtained by using all optical flow point pairs belonging to the object, all estimation results are sorted from small to large. Let the number of all estimation results be $M$, and take the average value of $(O_{i^{+}}^{k})_{m}$ at $0.1M, 0.2M, 0.3M$ position after ranking as the estimated value of object static probability $O_{i^{+}}^{k}$.


According to the calculation result of the static probability of the object and the real motion of the object, and taking into account that the negative effect of the dynamic point is generally greater than the positive effect of the increase of the static constraint when the camera pose is estimated, we set $O_{T h} = 0.9$, the object motion attributes are divided into high dynamic and low dynamic, which are provided to the subsequent methods as a priori information for processing with different strategies. The static probability of all keypoints in the box of the potential moving object is initialized to $O_{i^{+}}^{k}$, and the static probability of other keypoints is initialized to $1.0$.

\vspace{1mm}
\subsection{Static Probability of Keypoints in the First Stage}

\vspace{1mm}
\subsubsection{\bf{DBSCAN Density Clustering Algorithm}}

Compared with semantic segmentation methods, object detection technology has great advantages in real-time, but it can not provide accurate object mask. In the indoor dynamic SLAM scene, this problem leads to numerous static backgrounds in the boxes classified as people, and the false deletion of static keypoints will reduce the constraints of camera pose optimization and reduce the accuracy of camera pose estimation. We noticed that people as the foreground as a non-rigid body, his depth has a good continuity, and usually has a large fault with the background depth. To this end, we use the DBSCAN density clustering algorithm to distinguish between the foreground and background points of boxes classified as people. 



We adaptively determine $eps$ (the neighborhood radius of DBSCAN density clustering algorithm) and $minPts$ (the threshold of the number of samples in the neighborhood). After clustering, the one with the lowest average value of samples in cluster $\mathrm{C}=\left\{C_{1}, C_{2}, \cdots, C_{k}\right\}$ is taken as the foreground points of box.

After getting the DBSCAN clustering results, we adopt a soft strategy to further estimate the static probability of background points in the box of a potential moving object. Obviously, the static probability of background points must be greater than that of the object, and it is positively correlated with the static probability of the object. Specifies that the static probability of background points derived from the DBSCAN cluster is:

\vspace{-1mm}

\begin{equation}
K _{i}^{D k}=\left\{\begin{array}{c}
\frac{1-O_{T h}}{\left(O_{T h}\right)^{4}}\left(K_{i}^{k}\right)^{3}+1, O_{i^{+}}^{k} \leq O_{T h} \\
\frac{1}{K_{i}^{k}}, \quad O_{i^{+}}^{k}>O_{T h}
\end{array}\right.
    \label{eq6}
\end{equation}

Considering that the static probability estimation of keypoints has not been strictly calculated at each point, in other words, the static probability of the keypoints is coarse at present, and the camera pose estimation is vulnerable to dynamic points, we set the static probability of all foreground points in the box of high dynamic objects to $0$. 

\subsubsection{\bf{First Stage Pose Optimization}}

Update the static probability of keypoints:

\vspace{-3mm}

\begin{equation}
K_{i}^{k}=K_{i}^{k} \times K _{i}^{D k}
    \label{eq8}
\end{equation}

\vspace{-2mm}

When initializing the SLAM system, map points will be created. At this time, the static probability of map point $M_{i^{-}}^{k}$ will be initialized to the static probability of corresponding keypoint $K_{i}^{k}$. In the frame after initialization, $K_{i}^{k}$ and $M_{i^{-}}^{k}$ are used as weights to optimize the camera pose, and the camera pose estimation value $T_{k,w}$ in the first stage is obtained. Then, the static probability of $P_{i}^{k}$, which has a matching relation with the keypoints in $F_{k-1}$, is calculated precisely based on the projection constraints and the epipolar constraints.

\subsection{Static Probability of Keypoints in the Second Stage}

\subsubsection{\bf{Static Probability Based on the Projection Constraints}}

Convert the $P_{i^{*}}^{k-1}$ from the pixel coordinate to the camera coordinate:

\vspace{-2mm}

\begin{equation}
P_{i_{k-1}^{*}}^{k-1}=\frac{1}{K} Z_{i_{k-1}^{*}}^{k-1} \widetilde{P_{i_{u \nu}^{*}}^{k-1}}
    \label{eq9}
\end{equation}

Transform and project $P_{i_{k-1}^{*}}^{k-1}$ to $F_{k}$, and the Euclidean distance between the projection point and $P_{i}^{k}$ is:

\vspace{-4mm}

\begin{equation}
d_{i}^{T}=\left\|P_{i_{u v}}^{k}-\left.\left.\left|\frac{1}{\left|T_{k, k-1} \widetilde{P_{i_{k-1}^{*}}^{k-1}}\right|_{Z}} K\right| T_{k, k-1} \widetilde{P_{i_{k-1}^{*}}^{k-1}}\right|_{X Y Z}\right|_{u \nu}\right\|_{2} 
    \label{eq10}
\end{equation}

Where function $|\boldsymbol{P}|_{Z}$ represents the z-axis coordinate of point $\boldsymbol{P}$, and $|\boldsymbol{P}|_{XYZ}$ represents the non-homogeneous coordinate form of point $\boldsymbol{P}$.
On the premise that the camera pose $T_{k,w}$ is relatively accurate, the greater $d_{i}^{T}$, the greater the possibility that $P_{i}^{k}$ and $P_{i^{*}}^{k-1}$ are mismatched. Based on this principle, we design a static probability model based on the projection constraints. After sorting the $d_{i}^{T}$ of all keypoints outside the box of the dynamic object in $F_{k}$ from small to large, take $d_{i}^{T}$ at the truncated position of 0.8 as the adaptive threshold $D_{Th}^{T}$ of the projection error, and obtain the minimum value $d_{min}^{T}$ of $d_{i}^{T}$. We use the Sigmoid function form to measure the static probability of keypoints of the matching relationship in the box:

\begin{equation}
K_{i}^{T k}=\frac{1}{1+e^{\left(d_{i}^{T}-D_{T h}^{T}\right) \times} \frac{5}{D_{T h}^{T}-d_{\min }^{T}}}
    \label{eq11}
\end{equation}

For a pair of matching points, the satisfaction of the projection constraints is not only related to whether the corresponding spatial points strictly meet the static environment assumption, but also directly related to the number of constraints when solving the pose matrix and whether the pose matrix itself is correctly solved. Therefore, the statistical confidence $C_{s}^{T k}$ and calculation confidence $C_{c}^{T k}$ of the pose matrix are introduced:

\begin{equation}
C_{S}^{T k}=\frac{1}{1+e^{-N_{B A}+0.5 T h_{B A}}} 
    \label{eq12}
\end{equation}

\vspace{-1mm}

\begin{equation}
C_{C}^{T k}=1-\frac{\sum d_{i}^{T}}{N_{T} \times D_{T h}^{T}}
    \label{eq13}
\end{equation}

Where $N_{B A}$ is the number of interior points obtained by participating in the last camera pose solution, and threshold $T h_{B A}$ is the minimum number of interior points required to participate in the camera pose solution, $N_{T}$ and $\sum d_{i}^{T}$ respectively represent the number of all sample points and the sum of $d_{i}^{T}$ satisfying $d_{i}^{T}<D_{T h}^{T}$.

\vspace{1mm}

\subsubsection{\bf{Static Probability Based on the Epipolar Constraints}}

Based on the camera pose estimation $T_{k,w}$ in the first stage, a more accurate fundamental matrix  can be calculated:

\vspace{-1mm}
\begin{equation}
F_{k, k-1}=\mathrm{K}^{-\mathrm{T}}\left(t_{k, k-1}\right)^{\wedge} R_{k, k-1} \mathrm{~K}^{-1}
    \label{eq13}
\end{equation}

The pole line $l_{i}^{k}=\left[A_{i}^{k}, B_{i}^{k}, C_{i}^{k}\right]^{T}$ corresponding to $P_{i}^{k}$ is:

\vspace{-1mm}

\begin{equation}
l_{i}^{k}=F_{k, k-1} \widetilde{P_{i_{u v}^{*}}^{k-1}}
    \label{eq14}
\end{equation}

Then the polar error $d_{i}^{F}$ is:

\begin{equation}
d_{i}^{F}=\frac{\left|\left(\widetilde{P_{i_{\mathrm{uv}}}^{k}}\right)^{T} l_{i}^{k}\right|}{\sqrt{\left(A_{i}^{k}\right)^{2}+\left(B_{i}^{k}\right)^{2}}}
    \label{eq15}
\end{equation}

Similar to the projection constraints, we calculate static probability and confidence based on the epipolar constraints to obtain $K_{i}^{F k}$, the statistical confidence $C_{s}^{F k}$ and calculation confidence $C_{c}^{F k}$ of the fundamental matrix.





It should be noted that, as Eq.\ref{eq13} mentioned, the fundamental matrix can not be obtained when the camera translation is not large enough. Therefore, when the camera translation is less than the set threshold $t_{T h}$, skip the calculation of static probability and confidence based on the epipolar constraints, that is:

\vspace{-4mm}
\begin{equation}
K_{i}^{F k}=0, C_{S}^{F k}=C_{C}^{F k}=0 \quad \operatorname{s.t.} \left\|t_{k, k-1}\right\|_{2} \leq t_{T h}
    \label{eq19}
\end{equation}

\begin{figure*}[!b]
	\vspace{-1.5mm}
\centering
    \includegraphics[scale=0.485]{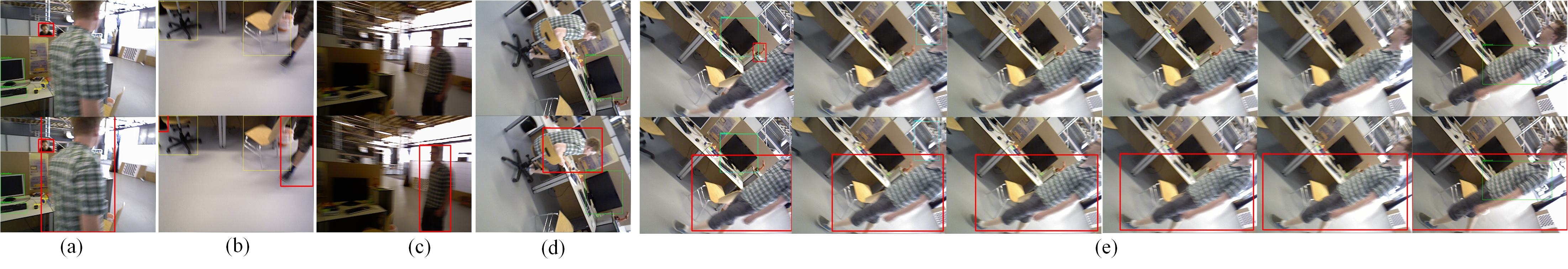}
 	\setlength{\abovecaptionskip}{-0.1cm} 
 	\caption{Missed detection and the results of missed detection compensation in the following cases: (a) The rapid  motion of the object. (b) The incomplete appearance of the object to be detected in the camera field of view. (c) The blurred image. (d) The singular angle of view caused by camera rotation. (e) Continuous frame miss detection.}
 	\label{missed detection compensation}
 \end{figure*}

\subsubsection{\bf{Second Stage Pose Optimization}}

After calculating the static probability of the keypoints based on the projection constraints and the epipolar constraints, we update the static probability of $P_{i}^{k}$ which matches the keypoints in $F_{k-1}$ for the second time. When the object is in high dynamics, the negative impact of dynamic points on camera pose estimation is generally greater than the positive impact of the increase in the number of static point constraints, which is just the opposite when the object is in low dynamics. This is because ORB-SLAM2 has certain outlier suppression strategies, which can suppress dynamic disturbances in low dynamics, but does not work in high dynamics. So, when $O_{i^{+}}^{k} \leq O_{T h}$,

\begin{equation}
K_{i}^{k}=\left\{\begin{array}{c}
K_{i}^{T k} \times K_{i}^{F k}, \left\|t_{k, k-1}\right\|_{2}>t_{T h} \\
K_{i}^{T k}, \quad \left\|t_{k, k-1}\right\|_{2} \leq t_{T h}
\end{array}\right.
    \label{eq21}
\end{equation}

when $O_{i^{+}}^{k}>O_{T h}$,

\vspace{-2mm}

\begin{equation}
K_{i}^{k}=\frac{K_{i}^{T k} \times C_{s}^{T k} C_{c}^{T k}}{C_{s}^{T k} C_{c}^{T k}+C_{s}^{F k} C_{c}^{F k}}+\frac{K_{i}^{F k} \times C_{s}^{F k} C_{c}^{F k}}{C_{s}^{T k} C_{c}^{T k}+C_{s}^{F k} C_{c}^{F k}}
    \label{eq22}
\end{equation}

After missed detection compensation, we use EKF and Hungarian algorithm to correlate the boxes of potential moving objects between adjacent frames. It is easy to know that if the association result of a box in $F_{k}$ is not found in $F_{k-1}$, even if there is a matching relationship between the foreground points in the box, it is generally a false matching, so let $K_{i}^{k}=0$ in this case. For $P_{i}^{k}$ that does not match the keypoints in $F_{k-1}$, according to the results of DBSCAN clustering, if $P_{i}^{k}$ belongs to the foreground points, let $K_{i}^{k}=0$, else let $K_{i}^{k}=M_{i^{-}}^{k}$. After the second estimation result of $K_{i}^{k}$ is obtained, $M_{i^{-}}^{k}$ is updated. When $M_{i^{-}}^{k} < 0.3$, delete the map point. Then $K_{i}^{k}$ and $M_{i^{-}}^{k}$ are used as weights to participate in the second stage of camera pose optimization. When there is a big difference between $K_{i}^{k}$ and $M_{i^{-}}^{k}$, it can be considered that $K_{i}^{k}$ and $M_{i^{-}}^{k}$ are mismatched and do not participate in optimization.

 \begin{figure*}[]
\centering
 	\includegraphics[scale=0.605]{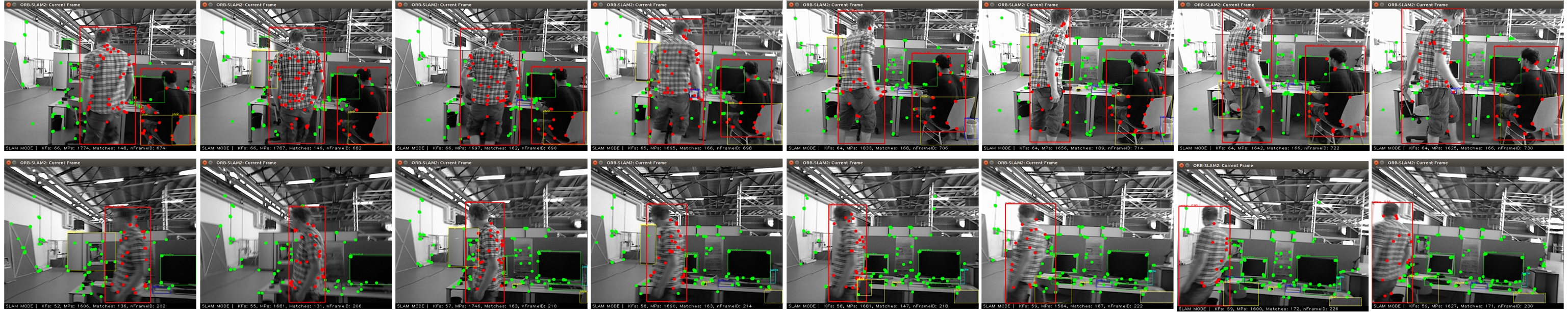}
 	\setlength{\abovecaptionskip}{-0.5cm} 
 	\caption{Effect of DBSCAN density clustering algorithm in two consecutive frames. The top set of images is taken every 8 frames, and the bottom set of images is taken every 4 frames. The images contain three common states of movement: sitting in a chair, slow motion and fast motion. After clustering, the foreground and background points are shown in red and green respectively.}
%
 	\label{DBSCAN}
 	\vspace{-3mm} 
 \end{figure*}

 \begin{figure*}[!b]
\centering
 	\includegraphics[scale=0.64]{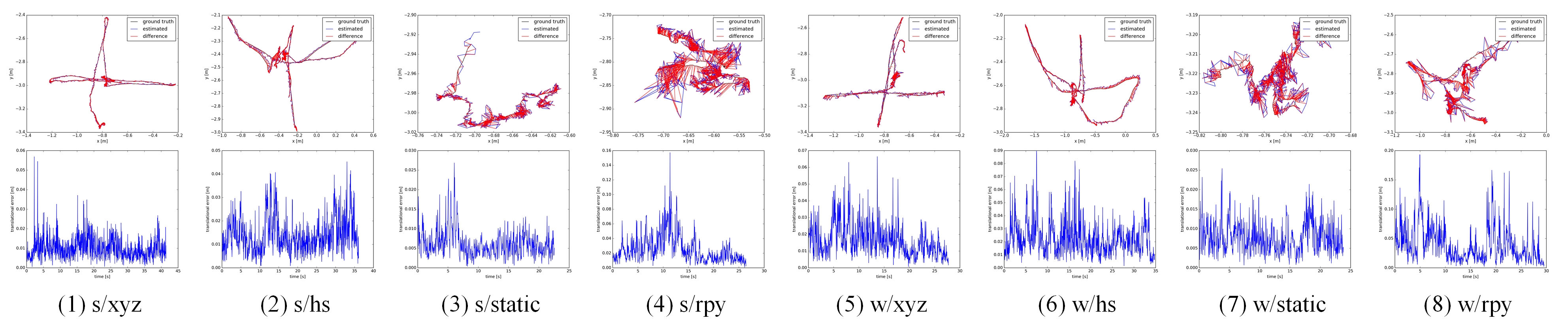}
 	\setlength{\abovecaptionskip}{-0.1cm} 
 	\caption{ATE and RPE from CFP-SLAM.}
 	\label{ATE and RPE}
 \end{figure*}


\begin{table*}[!b]
\begin{center}
\caption{RESULTS OF METRICS ABSOLUTE TRAJECTORY ERROR (ATE)}\label{ATE-lable}
\renewcommand\arraystretch{1.2}
\begin{tabular}{lcccccccccccccc}
\toprule
\specialrule{0em}{2pt}{1pt}
\multirow{2}{*}{Sequences} & \multicolumn{2}{c}{ORB-SLAM2}     & \multicolumn{2}{c}{Dyna-SLAM}    & \multicolumn{2}{c}{DS-SLAM} & \multicolumn{2}{c}{Blitz-SLAM} & \multicolumn{2}{c}{TRS} & \multicolumn{2}{c}{CFP-SLAM$^-$}    & \multicolumn{2}{c}{CFP-SLAM}          \\ 
\cline{2-15}
\specialrule{0em}{1pt}{2pt}
                           & RMSE            & S.D.            & RMSE            & S.D.           & RMSE      & S.D.            & RMSE           & S.D.          & RMSE        & S.D.      & RMSE            & S.D.            & RMSE            & S.D.            \\ \bottomrule 
fr3/s/xyz                        & \underline{0.0092}    & \underline{0.0047}    & 0.0127          & 0.0060          & /         & /               & 0.0148         & 0.0069        & 0.0117      & /         & 0.0129          & 0.0068          & \textbf{0.0090} & \textbf{0.0042} \\
fr3/s/half                       & 0.0192          & 0.0110          & 0.0186          & 0.0086          & /         & /               & 0.0160         & 0.0076        & 0.0172      & /         & \underline{0.0159}    & \underline{0.0072}    & \textbf{0.0147} & \textbf{0.0069} \\
fr3/s/static                     & 0.0087          & 0.0042          & /               & /               & 0.0065    & 0.0033          & /              & /             & /           & /         & \underline{0.0061}    & \underline{0.0029}    & \textbf{0.0053} & \textbf{0.0027} \\
fr3/s/rpy                        & \textbf{0.0195} & \textbf{0.0124} & /               & /               & /         & /               & /              & /             & /           & /         & \underline{0.0244}    & \underline{0.0175}    & 0.0253          & 0.0154          \\
fr3/w/xyz                        & 0.7214          & 0.2560          & 0.0164          & 0.0086          & 0.0247    & 0.0161          & 0.0153         & 0.0078        & 0.0194      & /         & \underline{0.0149}    & \underline{0.0077}    & \textbf{0.0141} & \textbf{0.0072} \\
fr3/w/half                       & 0.4667          & 0.2601          & 0.0296          & 0.0157          & 0.0303    & 0.0159          & 0.0256         & \underline{0.0126}  & 0.0290      & /         & \textbf{0.0235} & \textbf{0.0114} & \underline{0.0237}    & \textbf{0.0114} \\
fr3/w/static                     & 0.3872          & 0.1636          & \underline{0.0068}    & \underline{0.0032}    & 0.0081    & 0.0036    & 0.0102         & 0.0052        & 0.0111      & /       & 0.0069          & \underline{0.0032}   & \textbf{0.0066} & \textbf{0.0030}    \\
fr3/w/rpy                        & 0.7842          & 0.4005          & \textbf{0.0354} & \textbf{0.0190} & 0.4442    & 0.2350          & \underline{0.0356}   & \underline{0.0220}  & 0.0371      & /         & 0.0411          & 0.0257          & 0.0368          & 0.0230                            \\ 
\bottomrule
\end{tabular}
\end{center}
\end{table*}

\section{Experiments and Results} 

In this section, we test the performance of the proposed algorithm in 8 dynamic sequences of the TUM RGB-D dataset \cite{sturm2012benchmark}, including 4 low dynamic sequences (fr3/s for short) and 4 high dynamic sequences (fr3/w for short), and the camera includes 4 kinds of motion: static, xyz, halfsphere and rpy. The indicators used to evaluate the accuracy are the Absolute Trajectory Error (ATE) and the Relative Pose Error (RPE). ATE represents the global consistency of trajectory. RPE includes translation drift and rotation drift. The Root-Mean-Square-Error (RMSE) and Standard Deviation (S.D.) of both are used to represent the robustness and stability of the system\cite{fan2022blitz}. Firstly, we show the effect of missed detection compensation and DBSCAN clustering, then compare our method with some of the most advanced methods, then design a series of ablation experiments to test the impact of each module, and finally carry out real-time analysis. All the experiments are performed on a computer with Intel i7 CPU, 3060 GPU, and 16GB memory. 

\subsection{Missed Detection Compensation and DBSCAN Clustering}

In the dynamic SLAM scene, the motion of the object, the incomplete appearance of the object to be detected in the camera field of view, the blurred image and the singular angle of view caused by camera rotation all bring severe challenges to the object detection, very easy to cause miss detection, even will lead to continuous frame miss detection. Fig.\ref{missed detection compensation}(a)-(d) and Fig.\ref{missed detection compensation}(e) show the results of missed detection compensation of object detection in the above four cases and six consecutive frames, respectively. Fig.\ref{DBSCAN} shows the DBSCAN clustering results after missed detection compensation. We select two consecutive frames to show the clustering effect. The foreground points are marked with red and the background points are marked with green. The upper image group contains two people sitting on the chair and moving slowly respectively, and the people in the lower image group are in the fast walking state. It is worth noting from Fig.\ref{DBSCAN} that many 
keypoints are extracted from the edge of the person, which is generally the part with the highest dynamic attributes. However, semantic segmentation is difficult to accurately judge the boundary of objects\cite{fan2022blitz}, which leads to the misjudgment of dynamic and static attributes of keypoints. We use DBSCAN algorithm to cluster keypoints based on depth information, which can well avoid this problem. The experimental results fully show the effectiveness and robustness of the missed detection compensation algorithm and clustering algorithm.

\begin{table*}[]
\begin{center}
\caption{RESULTS OF METRIC TRANSLATIONAL DRIFT (RPE)}\label{T-RPE-lable}
\renewcommand\arraystretch{1.2}
\begin{tabular}{lcccccccccccccc}
\toprule
\specialrule{0em}{2pt}{1pt}
\multirow{2}{*}{Sequences} & \multicolumn{2}{c}{ORB-SLAM2}     & \multicolumn{2}{c}{Dyna-SLAM}    & \multicolumn{2}{c}{DS-SLAM} & \multicolumn{2}{c}{Blitz-SLAM} & \multicolumn{2}{c}{TRS} & \multicolumn{2}{c}{CFP-SLAM$^-$}    & \multicolumn{2}{c}{CFP-SLAM}          \\ 
\cline{2-15}
\specialrule{0em}{1pt}{2pt}
                           & RMSE            & S.D.            & RMSE                  & S.D.            & RMSE      & S.D.            & RMSE            & S.D.                  & RMSE           & S.D.   & RMSE            & S.D.            & RMSE            & S.D.                  \\ \bottomrule 
fr3/s/xyz                        & \underline{0.0117}    & \underline{0.0060}    & 0.0142                & 0.0073          & /            & /            & 0.0144          & 0.0071                & 0.0166         & /      & 0.0149          & 0.0081                & \textbf{0.0114} & \textbf{0.0055}       \\
fr3/s/half                       & 0.0231          & 0.0163          & 0.0239                & 0.0120          & /            & /            & \underline{0.0165}    & \textbf{0.0073}       & 0.0259         & /      & 0.0214    & 0.0099          & \textbf{0.0162} & \underline{0.0079} \\
fr3/s/static                     & 0.0090          & 0.0043          & /                     & /               & \underline{0.0078} & 0.0038       & /               & /                     & /              & /      & \underline{0.0078}    &  \textbf{0.0034} & \textbf{0.0072} & \underline{0.0035} \\
fr3/s/rpy                        & \textbf{0.0245} & \textbf{0.0144} & /                     & /               & /            & /            & /               & /                     & /              & /      & 0.0322    & 0.0217          & \underline{0.0316}    & \underline{0.0186}          \\
fr3/w/xyz                        & 0.3944          & 0.2964          & 0.0217                & 0.0119          & 0.0333       & 0.0229       & 0.0197          & \textbf{0.0096}       & 0.0234         & /      & \underline{0.0196}    & 0.0099          & \textbf{0.0190} & \underline{0.0097} \\
fr3/w/half                       & 0.3480          & 0.2859          & 0.0284                & 0.0149          & 0.0297       & 0.0152       & \textbf{0.0253} &  \textbf{0.0123} & 0.0423         & /      & 0.0274 & 0.0130       & \underline{0.0259}    & \underline{0.0128} \\
fr3/w/static                     & 0.2349          & 0.2151          & \textbf{0.0089} &  0.0044    & 0.0102       & 0.0048 & 0.0129          & 0.0069                & 0.0117         & /     & \underline{0.0092}    & \underline{0.0043}  & \textbf{0.0089} & \textbf{0.0040}                 \\
fr3/w/rpy                        & 0.4582          & 0.3447          & \textbf{0.0448}       & \textbf{0.0262} & 0.1503       & 0.1168       & 0.0473    & \underline{0.0283}          & \underline{0.0471}   & /      & 0.0540          & 0.0350                & 0.0500          & 0.0306              \\ 
\bottomrule
\end{tabular}
\end{center}
 	\vspace{-2mm}
\end{table*}

\begin{table*}[]
\begin{center}
\caption{RESULTS OF METRIC ROTATIONAL DRIFT (RPE)}\label{R-RPE-lable}
\renewcommand\arraystretch{1.2}
\begin{tabular}{lcccccccccccccc}
\toprule
\specialrule{0em}{2pt}{1pt}
\multirow{2}{*}{Sequences} & \multicolumn{2}{c}{ORB-SLAM2}     & \multicolumn{2}{c}{Dyna-SLAM}    & \multicolumn{2}{c}{DS-SLAM} & \multicolumn{2}{c}{Blitz-SLAM} & \multicolumn{2}{c}{TRS} & \multicolumn{2}{c}{CFP-SLAM$^-$}    & \multicolumn{2}{c}{CFP-SLAM}          \\ 
\cline{2-15}
\specialrule{0em}{1pt}{2pt}
                           & RMSE            & S.D.            & RMSE                  & S.D.            & RMSE      & S.D.            & RMSE            & S.D.                  & RMSE           & S.D.   & RMSE            & S.D.            & RMSE            & S.D.                  \\ \bottomrule 
fr3/s/xyz                  & \underline{0.4890}          & 0.2713          & 0.5042                & 0.2651                & /            & /            & 0.5024          & \textbf{0.2634}       & 0.5968         & /      & 0.5126          & 0.2793                & \textbf{0.4875}       & \underline{0.2640} \\
fr3/s/half                 & 0.6015                & 0.2924                & 0.7045                & 0.3488                & /            & /            & \underline{0.5981}    & \textbf{0.2739}       & 0.7891         & /      & 0.7697    & 0.3718          & \textbf{0.5917}       & \underline{0.2834} \\
fr3/s/static               & 0.2850                & 0.1241                & /                     & /                     & \underline{0.2735}  & 0.1215       & /               & /                     & /              & /      & 0.2749    & \underline{0.1192} & \textbf{0.2654}       & \textbf{0.1183} \\
fr3/s/rpy                  & \underline{0.7772} & \underline{0.3999} & /                     & /                     & /            & /            & /               & /                     & /              & /      & 0.8303   & 0.4653          & \textbf{0.7410} & \textbf{0.3665} \\
fr3/w/xyz                  & 7.7846                & 5.8335                & 0.6284                & 0.3848                & 0.8266       & 0.5826       & \underline{0.6132}    & \textbf{0.3348}       & 0.6368         & /      & 0.6204    & 0.3850          & \textbf{0.6023}       & \underline{0.3719} \\
fr3/w/half                 & 7.2138                & 5.8299                & \underline{0.7842}          & 0.4012                & 0.8142       & 0.4101       & 0.7879 & \underline{0.3751} & 0.9650         & /      & 0.7853 & 0.3821       & \textbf{0.7575} & \textbf{0.3743} \\
fr3/w/static               & 4.1856                & 3.8077                & 0.2612 & 0.1259          & 0.2690       & 0.1182 & 0.3038          & 0.1437                & 0.2872         & /    & \underline{0.2535}          & \underline{0.1130}   & \textbf{0.2527} & \textbf{0.1051}                \\
fr3/w/rpy                  & 8.8923                & 6.6658                & \textbf{0.9894}       & \underline{0.5701} & 3.0042       & 2.3065       & 1.0841    & 0.6668          & 1.0587   & /      & \underline{1.0521}    & \textbf{0.5577}       & 1.1084                & 0.6722                             \\ 
\bottomrule
\end{tabular}
\end{center}
 	\vspace{-2mm}
\end{table*}

\begin{table*}[h]
\begin{center}
\caption{RESULTS OF METRICS ABSOLUTE TRAJECTORY ERROR (ATE) WITH DIFFERENT CONFIGURATIONS}\label{Ab-ATE-lable}
\renewcommand\arraystretch{1.2}
\begin{tabular}{lcccccccccccc}
\toprule
\specialrule{0em}{2pt}{1pt}
\multirow{2}{*}{Sequences} & \multicolumn{2}{c}{CFP-SLAM}     & \multicolumn{2}{c}{CFP-SLAM$^{-}$}    & \multicolumn{2}{c}{W/O-MDC} & \multicolumn{2}{c}{W/O-DBS} & \multicolumn{2}{c}{W/O-KSP} & \multicolumn{2}{c}{Only-YOLO}            \\ 
\cline{2-13}
\specialrule{0em}{1pt}{2pt}
                           & RMSE            & S.D.            & RMSE            & S.D.           & RMSE      & S.D.            & RMSE           & S.D.          & RMSE        & S.D.      & RMSE            & S.D.                        \\ \bottomrule 
fr3/s/xyz                  & \textbf{0.0090} & \textbf{0.0042} & 0.0129                & 0.0068                & \underline{0.0123}    & 0.0066          & 0.0130          & \underline{0.0060} & 0.0142       & 0.0063          & 0.0174          & 0.0079                \\
fr3/s/half                 & \textbf{0.0147}       & \textbf{0.0069}       & 0.0159                & \underline{0.0072}          & \underline{0.0150}    & 0.0074          & 0.0305    & 0.0179       & 0.0201       & 0.0089          & 0.0281    & 0.0158          \\
fr3/s/static               & \textbf{0.0053}       & \underline{0.0027}          & 0.0061                & 0.0029                & \underline{0.0055}    & \textbf{0.0025} & 0.0064          & 0.0030                & 0.0062       & 0.0030          & 0.0064    & \underline{0.0027} \\
fr3/s/rpy                  & 0.0253 & \underline{0.0154} & \underline{0.0244}          & 0.0175                & \textbf{0.0237} & \textbf{0.0149} & 0.0297          & 0.0205                & 0.0287       & 0.0195          & 0.0460    & 0.0332          \\
fr3/w/xyz                  & \textbf{0.0141}       & \textbf{0.0072}       & \underline{ 0.0149}          & 0.0077          & 0.0158          & 0.0079          & 0.0159    & 0.0081       & 0.0154       & \underline{0.0076}          & 0.0165    & 0.0082          \\
fr3/w/half                 & \underline{0.0237}          & \textbf{0.0114}       & \textbf{0.0235} & \textbf{0.0114}       & 0.0258          & \underline{0.0134}    & 0.0274 & 0.0137 & 0.0307       & 0.0151          & 0.0310 & 0.0165       \\
fr3/w/static                               & \textbf{0.0066} & \textbf{0.0030}  & \underline{0.0069}          & 0.0032  & 0.0070          & \underline{0.0031}    & 0.0078          & 0.0033                & 0.0076       & 0.0033          & 0.0073 & 0.0032       \\
fr3/w/rpy                  & \textbf{0.0368}       & \underline{0.0230}          & 0.0411       & 0.0257 & 0.1910          & 0.1594          & 0.0749    & 0.0536          & \underline{0.0405} & \textbf{0.0211} & 0.0456    & 0.0312           \\ 
\bottomrule
\end{tabular}
\end{center}
\vspace{-7mm}
\end{table*}


\subsection{Comparison with State-of-the-arts}

We contrast with ORB-SLAM2 \cite{mur2017orb} and forth most advanced dynamic SLAM methods, including DS-SLAM \cite{yu2018ds}, Dyna-SLAM \cite{bescos2018dynaslam}, Blitz-SLAM \cite{fan2022blitz} and TRS \cite{ji2021towards}. Like our method, these algorithms are all improved based on ORB-SLAM2. Without calculating the static probability of the object, we provide a lower performance version of the algorithm in this paper with higher real-time performance, which is called CFP-SLAM$^{-}$. The quantitative comparison results are shown in Tables \ref{ATE-lable}, \ref{T-RPE-lable} and \ref{R-RPE-lable}, in which the best results are highlighted in bold and the second-best are underlined. The data of DS-SLAM, Dyna-SLAM, Blitz-SLAM and TRS comes from the source literature, $/$ indicates that the corresponding data is not provided in the source literature. The experimental results show that, unlike other dynamic SLAM algorithms, which only have advantages over ORB-SLAM2 in high dynamic scenarios, this algorithm can achieve almost the best results in high dynamic and low dynamic scenarios. Even the low-performance version we provide shows better performance than other algorithms. In rpy sequences, on the one hand, the epipolar constraints cannot be used, on the other hand, the large change of camera angle leads to insufficient feature matching, so our method performs slightly worse. The ATE and RPE plots of our algorithm on 8 sequences are shown in Fig.\ref{ATE and RPE}.


\subsection{Ablation Experiment}

In order to prove the function of each module of our algorithm, We design a series of ablation experiments, and the experimental results are shown in Table \ref{Ab-ATE-lable}. Among them, CFP-SLAM: The algorithm of this paper; CFP-SLAM$^{-}$: Do not use static probability of objects; W/O-MDC: Without missed detection compensation; W/O-DBS: Without DBSCAN clustering; W/O-KSP: Without the static probability of keypoints, that is, all the foreground points after missed detection compensation and DBSCAN clustering are directly eliminated; Only-YOLO: Directly eliminate all keypoints in the box with human category. 


The experimental results show that CFP-SLAM$^{-}$ shows worse performance in low dynamic scenes, because we cannot distinguish between high dynamic objects and low dynamic objects, so all objects are processed according to high dynamic. W/O-MDC is almost unaffected in low dynamic scenes, but the performance is very poor in high dynamic scenes, especially in w/rpy, when the camera and objects are moving violently. In fact, the tracking is often lost in w/xyz, w/half and w/rpy because of missed detection. W/O-DBS and W/O-KSP show general performance in all sequences, which illustrates the effectiveness of DBSCAN clustering and the limitation of dealing with non-rigid bodies with partial motion as a whole, respectively. Only-YOLO encounters difficulties in initialization due to insufficient features in almost all sequences, and tracking is lost in some sequences.

\subsection{Real-time Analysis}

Real-time performance is one of the important evaluation indexes of SLAM system. We test the average running time of each module, as shown in Table \ref{Real-time analysis}. EKF represents the missed detection compensation and data association of boxes module, OSP represents the static probability calculation module of objects, and KSP represents the static probability calculation module of keypoints based on the epipolar constraints and the projection constraints. Semantic threads based on YOLOv5s run in parallel with ORB feature extraction. The results show that the average processing time per frame for the main threads of CFP-SLAM and CFP-SLAM$^{-}$ is 42.7 ms and 24.77 ms, that is, the running speed reaches 23 Fps and 40 Fps respectively. Compared with the SLAM system based on semantic segmentation, it can better meet the real-time requirements while ensure the accuracy.

\begin{table}[h]
\begin{center}
\caption{THE AVERAGE RUNNING TIME OF EACH MODULE.}\label{Real-time analysis}
\setlength{\belowcaptionskip}{-0.5cm}   
\setlength{\tabcolsep}{1.6mm}{
\begin{tabular}{|l|c|c|c|c|c|c|}
\hline
\multicolumn{1}{|c|}{Methods} & YOLO & EKF  & OSP   & \multicolumn{1}{l|}{DBSCAN} & KSP  & Tracking \\ \hline
CFP-SLAM                      & 12.44   & 0.07 & 17.93 & 1.76                        & 3.66 & 42.7     \\ \hline
CFP-SLAM-                     & 12.44   & 0.07 & /     & 1.76                        & 3.66 & 24.77    \\ \hline
\end{tabular}
}
\end{center}
\vspace{-5mm}
\end{table}

\section{Conclusion} 

In this paper, we propose a dynamic scene-oriented visual SLAM algorithm based on YOLOv5s and coarse-to-fine static probability. After missed detection compensation and keypoints clustering, the static probabilities of objects, keypoints and map points are calculated and updated as weights to participate in pose optimization. Extensive evaluation shows that our algorithm achieves the highest accuracy of localization in almost all low dynamic and high dynamic scenes, and has quite high real-time performance. In the future, we intend to build a lightweight plane and object map containing only static environment for robot navigation and augmented reality.



\bibliographystyle{IEEEtran}
\bibliography{CFP-SLAM}



\end{document}